\def\year{2018}\relax
\documentclass[letterpaper]{article} 
\usepackage{aaai18}  
\usepackage{times}  
\usepackage{helvet}  
\usepackage{courier}  
\usepackage{url}  
\usepackage{graphicx}  

\usepackage{amsfonts}
\usepackage{amsmath}
\usepackage{graphicx}
\usepackage{caption}
\usepackage{subcaption}
\usepackage{algorithm}
\usepackage{algpseudocode}
\usepackage{multirow}
\usepackage{array}

\begin{document}
%
\title{Multi-Task Label Embedding for Text Classification}
\author{Honglun Zhang$^{1}$, Liqiang Xiao$^{1}$, Wenqing Chen$^{1}$, Yongkun Wang$^{2}$, Yaohui Jin$^{1,2}$ \\
$^{1}$State Key Lab of Advanced Optical Communication System and Network \\
$^{2}$Network and Information Center \\ 
Shanghai Jiao Tong University \\
\{ykw\}@sjtu.edu.cn}


\maketitle
\begin{abstract}
Multi-task learning in text classification leverages implicit correlations among related tasks to extract common features and yield performance gains. However, most previous works treat labels of each task as independent and meaningless one-hot vectors, which cause a loss of potential information and makes it difficult for these models to jointly learn three or more tasks. In this paper, we propose Multi-Task Label Embedding to convert labels in text classification into semantic vectors, thereby turning the original tasks into vector matching tasks. We implement unsupervised, supervised and semi-supervised models of Multi-Task Label Embedding, all utilizing semantic correlations among tasks and making it particularly convenient to scale and transfer as more tasks are involved. Extensive experiments on five benchmark datasets for text classification show that our models can effectively improve performances of related tasks with semantic representations of labels and additional information from each other.
\end{abstract}

\section{Introduction}

Text classification is a common \textit{Natural Language Processing} task that tries to infer the most appropriate label for a given sentence or document, for example, sentiment analysis, topic classification and so on. With the developments and prosperities of \textit{Deep Learning}~\cite{DBLP:journals/pami/BengioCV13}, many neural network based models have been exploited by a large body of literature and achieved inspiring performance gains on various text classification tasks. These models are robust at feature engineering and can represent word sequences as fix-length vectors with rich semantic information, which are notably ideal for subsequent NLP tasks.

Due to numerous parameters to train, neural network based models rely heavily on adequate amounts of annotated corpora, which can not always be met as constructions of large-scale high-quality labeled datasets are extremely time-consuming and labor-intensive. \textit{Multi-Task Learning} solves this problem by jointly training multiple related tasks and leveraging potential correlations among them to increase corpora size implicitly, extract common features and yield classification improvements. Inspired by~\cite{DBLP:journals/ml/Caruana97}, there are lots of works dedicated for multi-task learning with neural network based models~\cite{DBLP:conf/icml/CollobertW08,DBLP:conf/naacl/LiuGHDDW15,DBLP:conf/emnlp/LiuQH16,DBLP:conf/ijcai/LiuQH16,ijcai2017-473}. These models usually contain a pre-trained lookup layer that map words into dense, low-dimension and real-value vectors with semantic implications, which is known as \textit{Word Embedding}~\cite{DBLP:conf/nips/MikolovSCCD13}, and utilize some lower layers to capture common features that are further fed to follow-up task-specific layers. However, most existing models have the following three disadvantages: 

\begin{itemize}
\item Lack of \textbf{Label Information}. Labels of each task are represented by independent and meaningless one-hot vectors, for example, \textit{positive} and \textit{negative} in sentiment analysis encoded as $[1,0]$ and $[0,1]$, which may cause a loss of potential label information.  
\item Incapable of \textbf{Scaling}. Network structures are elaborately designed to model various correlations for multi-task learning, but most of them are structurally fixed and can only deal with interactions between two tasks, namely pair-wise interactions. When new tasks are introduced, the network structures have to be modified and the whole networks have to be trained again.
\item Incapable of \textbf{Transferring}. For human beings, we can handle a completely new task without any more efforts after learning with several related tasks, which can be concluded as the capability of \textit{Transfer Learning}~\cite{DBLP:conf/kdd/LingDXYY08}. As discussed above, the network structures of most previous models are fixed, thus not compatible with and failing to tackle new tasks.
\end{itemize}

In this paper, we proposed \textbf{Multi-Task Label Embedding}~(\textbf{MTLE}) to map labels of each task into semantic vectors as well, similar to how Word Embedding represents the word sequences, thereby converting the original text classification tasks into vector matching tasks. Based on MTLE, we implement unsupervised, supervised and semi-supervised multi-task learning models for text classification, all utilizing semantic correlations among tasks and effectively solving the problems of scaling and transferring when new tasks are involved.

We conduct extensive experiments on five benchmark datasets for text classification. Compared to learning separately, jointly learning multiple related tasks based on MTLE demonstrates significant performance gains for each task.

Our contributions are four-folds:

\begin{itemize}
\item Our models efficiently leverage potential label information of each task by mapping labels into dense, low-dimension and real-value vectors with semantic implications.
\item It is particularly convenient for our models to scale when new tasks are involved. The network structures need no modifications and only data from the new tasks require training.
\item After training on several related tasks, our models can also naturally transfer to deal with completely new tasks without any additional training, while still achieving appreciable performances.
\item We consider different scenarios of multi-task learning and demonstrate strong results on several benchmark datasets for text classification. Our models outperform most state-of-the-art baselines.
\end{itemize}

\section{Problem Statements}

\subsection{Single-Task Learning}

In a supervised text classification task, the input is a word sequence denoted by $x=\{x_1,x_2,...,x_T\}$ and the output is the class label $y$ or the one-hot representation $\mathbf{y}$. A pre-trained lookup layer is used to get the embedding vector $\mathbf{x}_t\in\mathbb{R}^d$ for each word $x_t$. A text classification model $f$ is trained to produce the predicted distribution $\hat{\mathbf{y}}$ for each $\mathbf{x}=\{\mathbf{x}_1,\mathbf{x}_2,...,\mathbf{x}_T\}$.
\begin{equation}\tag{$1$}\label{eq:1}
f(\mathbf{x}_1,\mathbf{x}_2,...,\mathbf{x}_T)=\hat{\mathbf{y}}
\end{equation}
and the training objective is to minimize the total cross-entropy over all samples.
\begin{equation}\tag{$2$}\label{eq:2}
l=-\sum_{i=1}^{N}\sum_{j=1}^{C}y_{ij}\log{\hat{y}_{ij}}
\end{equation}
where $N$ denotes the number of training samples and $C$ is the class number.

\subsection{Multi-Task Learning}

Given $K$ supervised text classification tasks, $T_1,T_2,...,T_K$, a multi-task learning model $F$ is trained to transform each $\mathbf{x}^{(k)}$ from $T_k$ into multiple predicted distributions $\{\hat{\mathbf{y}}^{(1)},\hat{\mathbf{y}}^{(2)},...,\hat{\mathbf{y}}^{(K)}\}$.

\begin{equation}\tag{$3$}\label{eq:3}
F(\mathbf{x}_1^{(k)}, \mathbf{x}_2^{(k)},...,\mathbf{x}_T^{(k)})=\{\hat{\mathbf{y}}^{(1)},\hat{\mathbf{y}}^{(2)},...,\hat{\mathbf{y}}^{(K)}\}
\end{equation}
where only $\hat{\mathbf{y}}^{(k)}$ is used for loss computation. The overall training loss is a weighted linear combination of costs for each task.
\begin{equation}\tag{$4$}\label{eq:4}
L=-\sum_{k=1}^{K}\lambda_k\sum_{i=1}^{N_k}\sum_{j=1}^{C_k}y_{ij}^{(k)}\log{\hat{y}_{ij}^{(k)}}
\end{equation}
where $\lambda_k$, $N_k$ and $C_k$ denote the linear weight, the number of samples and the class number for each task $T_k$ respectively.

\subsection{Three Perspectives of Multi-Task Learning}

Text classification tasks can differ in characteristics of the input word sequence $\mathbf{x}$ or the output label $\mathbf{y}$. There are lots of benchmark datasets for text classification and three different perspectives of multi-task learning can be concluded.

\begin{itemize}
\item \textbf{Multi-Cardinality} Tasks are similar apart from cardinalities, for example, movie review datasets with different average sequence lengths and class numbers. 
\item \textbf{Multi-Domain} Tasks are different in domains of corpora, for example, product review datasets on books, DVDs, electronics and kitchen appliances. 
\item \textbf{Multi-Objective} Tasks are targeted for different objectives, for example, sentiment analysis, topic classification and question type judgment.
\end{itemize}

The simplest multi-task learning scenario is that all tasks share the same cardinality, domain and objective, while just come from different sources. On the contrary, when tasks vary in cardinality, domain and even objective, the correlations and interactions among them can be quite complicated and implicit. When implementing multi-task learning, both the model used and the tasks involved have significant influences on the ideal performance gains for each task. We will further investigate the scaling and transferring capabilities of MTLE on different scenarios in the Experiment section.

\section{Methodology}

Neural network based models have obtained substantial interests in many NLP tasks for their capabilities to represent variable-length words sequences as fix-length vectors, for example, \textit{Neural Bag-of-Words} (NBOW), \textit{Recurrent Neural Networks} (RNN), \textit{Recursive Neural Networks} (RecNN) and \textit{Convolutional Neural Network} (CNN). These models mostly first map sequences of words, n-grams or other semantic units into embedding representations with a pre-trained lookup layer, then comprehend the vector sequences with neural networks of different structures and mechanisms, finally utilize a softmax layer to predict categorical distribution for specific text classification tasks. For RNN, input vectors are absorbed one by one in a recurrent manner, which resembles the way human beings understand texts and makes RNN notably suitable for NLP tasks.

\subsection{Recurrent Neural Network}

RNN maintains a internal hidden state vector $\mathbf{h}_t$ that is recurrently updated by a transition function $f$. At each time step $t$, the hidden state $\mathbf{h}_t$ is updated according to the current input vector $\mathbf{x}_t$ and the previous hidden state $\mathbf{h}_{t-1}$.
\begin{equation}\tag{$5$}\label{eq:5}
\mathbf{h}_t=\left\{
\begin{array}{ll}
0 & t=0\\
f(\mathbf{h}_{t-1},\mathbf{x}_t) & \mbox{otherwise}
\end{array}
\right.
\end{equation}
where $f$ is usually a composition of an element-wise nonlinearity with an affine transformation of both $\mathbf{x}_t$ and $\mathbf{h}_{t-1}$. In this way, RNN can accept a word sequence of arbitrary length and produce a fix-length vector, which is fed to a softmax layer for text classification or other NLP tasks. However, gradient of $f$ may grow or decay exponentially over long sequences during training, namely the \textit{gradient exploding} or \textit{vanishing} problems, which hinder RNN from effectively learning long-term dependencies and correlations.

\cite{DBLP:journals/neco/HochreiterS97} proposed \textit{Long Short-Term Memory Network} (LSTM) to solve the above problems. Besides the internal hidden state $\mathbf{h}_t$, LSTM also maintains an internal memory cell and three gating mechanisms. While there are numerous variants of the standard LSTM, in this paper we follow the implementation of~\cite{DBLP:journals/corr/Graves13}. At each time step $t$, states of the LSTM can be fully described by five vectors in $\mathbb{R}^m$, an \textit{input gate} $\mathbf{i}_t$, a \textit{forget gate} $\mathbf{f}_t$, an \textit{output gate} $\mathbf{o}_t$, the \textit{hidden state} $\mathbf{h}_t$ and the \textit{memory cell} $\mathbf{c}_t$, which adhere to the following transition equations.
\begin{align*}
&\mathbf{i}_t=\sigma(\mathbf{W}_i\mathbf{x}_t+\mathbf{U}_i\mathbf{h}_{t-1}+\mathbf{V}_i\mathbf{c}_{t-1}+\mathbf{b}_i)\tag{$6$}\label{eq:6}\\
&\mathbf{f}_t=\sigma(\mathbf{W}_f\mathbf{x}_t+\mathbf{U}_f\mathbf{h}_{t-1}+\mathbf{V}_f\mathbf{c}_{t-1}+\mathbf{b}_f)\tag{$7$}\label{eq:7}\\
&\mathbf{o}_t=\sigma(\mathbf{W}_o\mathbf{x}_t+\mathbf{U}_o\mathbf{h}_{t-1}+\mathbf{V}_o\mathbf{c}_{t-1}+\mathbf{b}_o)\tag{$8$}\label{eq:8}\\
&\tilde{\mathbf{c}}_t=\tanh(\mathbf{W}_c\mathbf{x}_t+\mathbf{U}_c\mathbf{h}_{t-1})\tag{$9$}\label{eq:9}\\
&\mathbf{c}_t=\mathbf{f}_t\odot\mathbf{c}_{t-1}+\mathbf{i}_t\odot\tilde{\mathbf{c}}_t\tag{$10$}\label{eq:10}\\
&\mathbf{h}_t=\mathbf{o}_t\odot\tanh(\mathbf{c}_t)\tag{$11$}\label{eq:11}
\end{align*}
where $\mathbf{x}_t$ is the current input, $\sigma$ denotes logistic sigmoid function and $\odot$ denotes element-wise multiplication. By strictly controlling how to accept $\mathbf{x}_t$ and the portions of $\mathbf{c}_t$ to update, forget and expose at each time step, LSTM can better understand long-term dependencies according to the labels of the whole sequences.

\subsection{Multi-Task Label Embedding}

Labels of text classification tasks are made up of word sequences as well, for example, \textit{positive} and \textit{negative} in binary sentiment classification, \textit{very positive}, \textit{positive}, \textit{neutral}, \textit{negative} and \textit{very negative} in 5-categorical sentiment classification. Inspired by Word Embedding, we propose \textbf{Multi-Task Label Embedding}~(\textbf{MTLE}) to convert labels of each task into dense, low-dimension and real-value vectors with semantic implications, thereby disclosing potential intra-task and inter-task label correlations. 

\begin{figure}[!ht]
\centering
\includegraphics[width=0.45\textwidth]{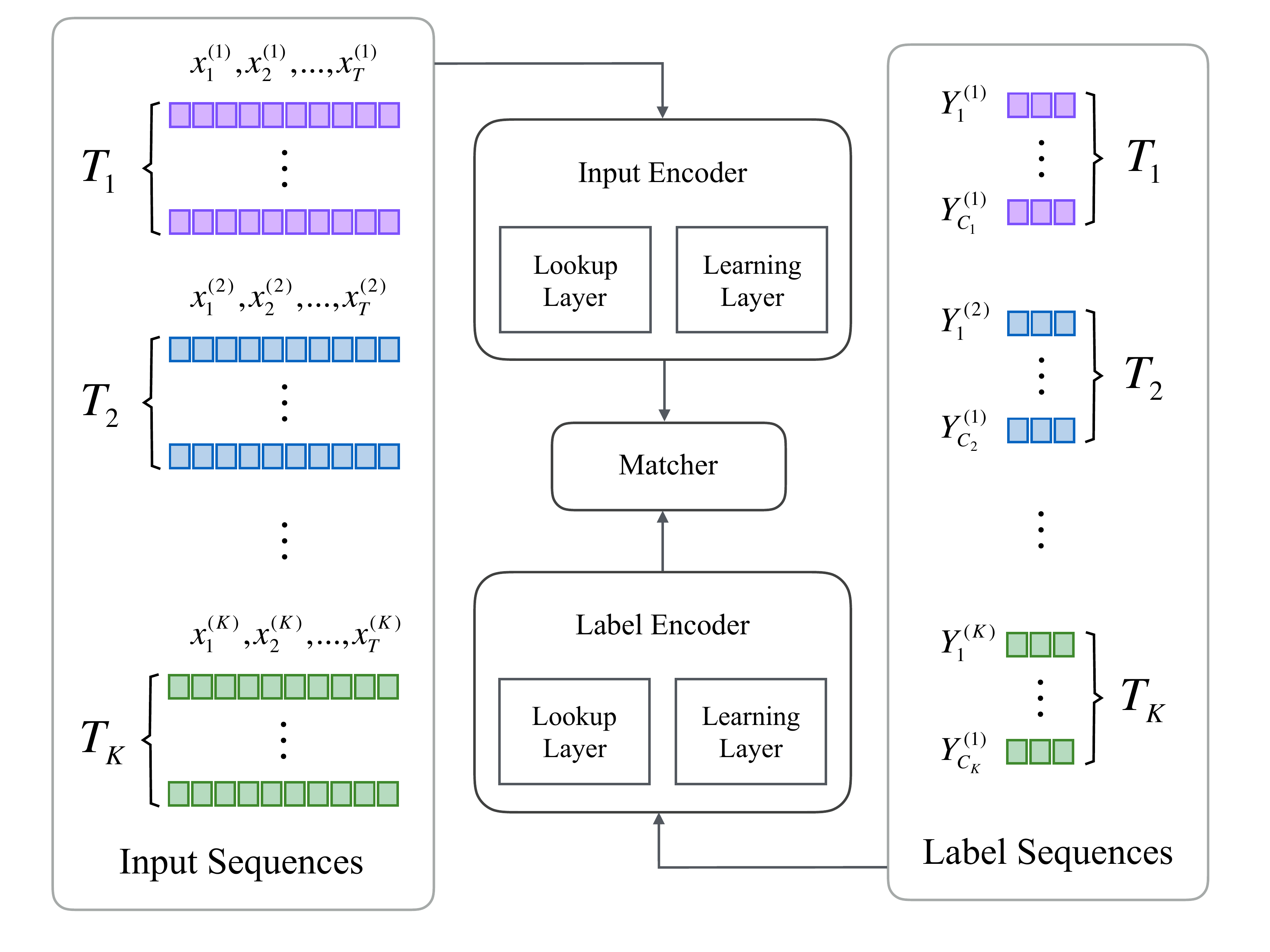}
\caption{General idea of MTLE for text classification}\label{architecture}
\end{figure}

Figure \ref{architecture} illustrates the general idea of MTLE for text classification, which mainly consists of three parts, the \textbf{Input Encoder}, the \textbf{Label Encoder} and the \textbf{Matcher}.

In the Input Encoder, each input sequence $x^{(k)}=\{x_1^{(k)},x_2^{(k)},...,x_T^{(k)}\}$ from $T_k$ is transformed into its embedding representation $\mathbf{x}^{(k)}=\{\mathbf{x}_1^{(k)},\mathbf{x}_2^{(k)},...,\mathbf{x}_T^{(k)}\}$ by the Lookup Layer ($Lu_{I}$). The Learning Layer ($Le_{I}$) is applied to recurrently comprehend $\mathbf{x}^{(k)}$ and generate a fix-length vector $\mathbf{X}^{(k)}$, which can be regarded as an overall representation of the original input sequence $x^{(k)}$.

In the Label Encoder, labels of each task are mapped and learned to produce fix-length representations as well. There are $C_k$ labels in $T_k$, namely $y_1^{(k)},y_2^{(k)},...,y_{C_k}^{(k)}$, where $y_{j}^{(k)}(1\leq j \leq C_k)$ is also a word sequence, for example, \textit{very positive}, and is mapped into the vector sequence $\mathbf{y}_{j}^{(k)}$ by the Lookup Layer ($Lu_{L}$). The Learning Layer ($Le_{L}$) further absorb $\mathbf{y}_{j}^{(k)}$ to generate a fix-length vector $\mathbf{Y}_j^{(k)}$, which can be concluded as an overall semantic representation of the original label $y_{j}^{(k)}$.

In order to achieve the classification task for a sample $x^{(k)}$ from $T_k$, the Matcher obtains the corresponding $\mathbf{X}^{(k)}$ from the Input Encoder, all $\mathbf{Y}_j^{(k)}(1\leq j \leq C_k)$ from the Label Encoder, and then conducts vector matching to select the most appropriate class label.

Based on the idea of MTLE, we implement unsupervised, supervised and semi-supervised models to investigate and explore different possibilities of multi-task learning in text classification.

\subsection{Model-\uppercase\expandafter{\romannumeral1}: Unsupervised}

Suppose that for each task $T_k$, we only have $N_k$ input sequences and $C_k$ classification labels, but lack the specific annotations for each input sequence and its corresponding label. In this case, we can only implement MTLE in an unsupervised manner.

Word Embedding~\cite{DBLP:conf/nips/MikolovSCCD13} leverages contextual features of words and trains them into semantic vectors so that words sharing synonymous meanings result in vectors of similar values. In the unsupervised model, we utilize all available input sequences and classification labels as the whole corpora and train a embedding model $E_{unsup}$~\cite{DBLP:journals/corr/abs-1301-3781} that covers contextual features of different tasks. The embedding model will be employed as both $Lu_{I}$ and $Lu_{L}$.

We achieve $Le_{I}$ and $Le_{L}$ simply by summing up vectors in a sequence and calculating the average, since we don't have any supervised annotations. After obtaining $\mathbf{X}^{(k)}$ for each input sample and all $\mathbf{Y}_j^{(k)}$ for a certain task $T_k$, we apply unsupervised vector matching methods $D(\mathbf{X}^{(k)},\mathbf{Y}_j^{(k)})$, for example, \textit{Cosine Similarity} or \textit{$L_2$ Distance}, to select the most appropriate $\mathbf{Y}_j^{(k)}$ for each $\mathbf{X}^{(k)}$.

In conclusion, the unsupervised model of MTLE exploits contextual and semantic information of both the input sequences and the classification labels. Model-\uppercase\expandafter{\romannumeral1} may fail to achieve adequately satisfactory performances due to employments of so many unsupervised methods, but can still provide some useful insights when no annotations are available at all.

\subsection{Model-\uppercase\expandafter{\romannumeral2}: Supervised}

Given the specific annotations for each input sequence and its corresponding label, we can better train the Input Encoder and the Label Encoder in a supervised manner. 

The $Lu_{I}$ and the $Lu_{L}$ are both fully-connection layers with the weights $\mathbf{W}_I$ and $\mathbf{W}_L$ of $\left|V \right| \times d$ matrixes, where $\left|V \right|$ denotes the vocabulary size and $d$ is the embedding size. We can utilize the $E_{unsup}$ obtained in Model-\uppercase\expandafter{\romannumeral1} or other pre-trained lookup tables to initialize $\mathbf{W}_I,\mathbf{W}_L$ and further tune their weights during training.

The $Le_{I}$ and the $Le_{L}$ should be trainable models that can transform a vector sequence of arbitrary lengths into a fix-length vector. We apply the implementation of~\cite{DBLP:journals/corr/Graves13} and denote them by $LSTM_{I}$ and $LSTM_{L}$ with hidden size $m$. We can also try some more complicated but effective sequence learning models, but in this paper we mainly focus on the idea and effects of MTLE, so we just choose a common one for implementation and spend more efforts on explorations of MTLE.

\begin{figure}[!ht]
\centering
\includegraphics[width=0.45\textwidth]{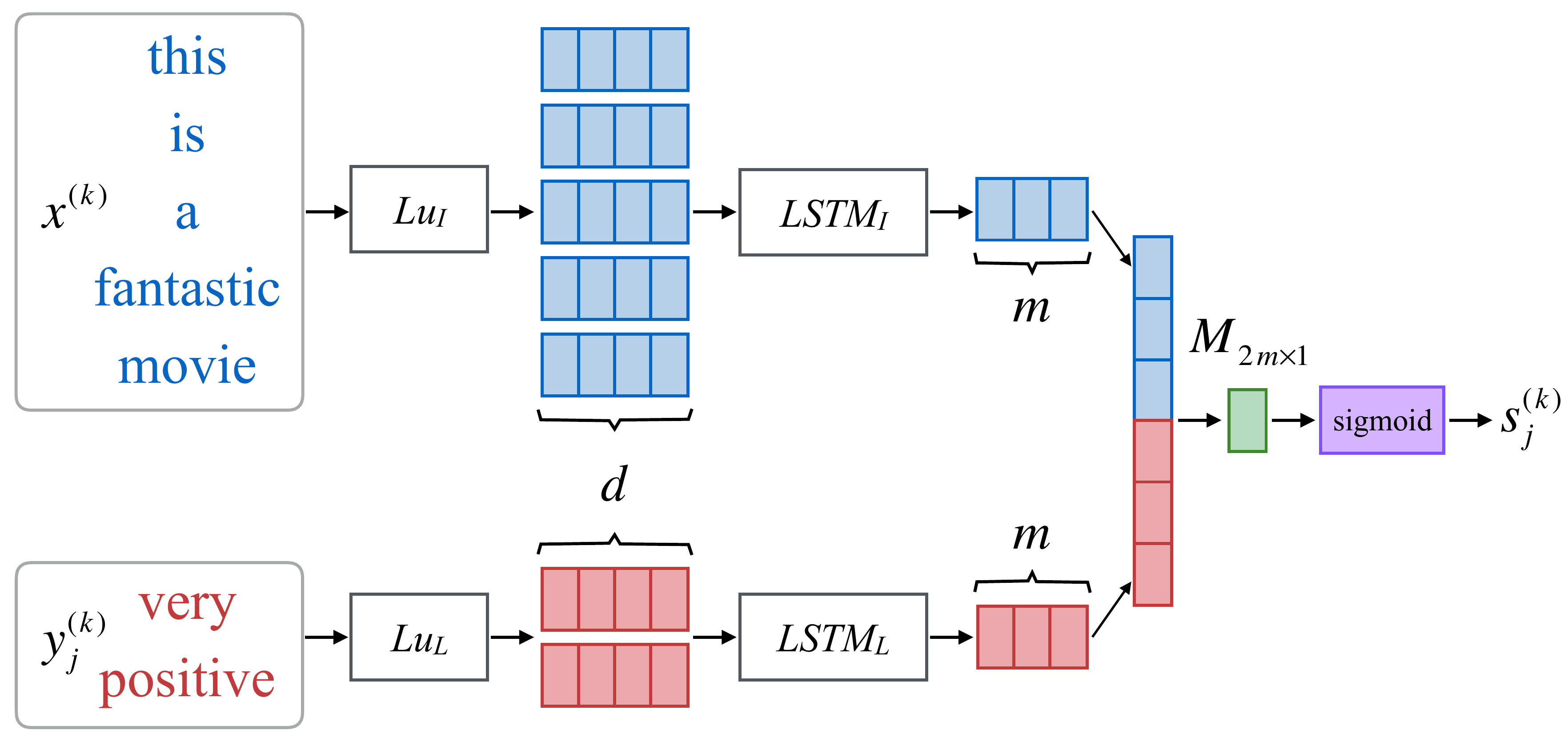}
\caption{Supervised model for MTLE}\label{supervised}
\end{figure}

We utilize another fully-connection layer of size $2m \times 1$, denoted by $M_{2m\times 1}$, to achieve the Matcher, which accepts outputs from the $Le_{I}$ and the $Le_{L}$ to produce a score of matching. Given the matching scores of each label, we implement the idea of cross-entropy and calculate the loss function for a sample $x^{(k)}$ from $T_k$ as follows.

\begin{align*}
&\mathbf{X}^{(k)}=LSTM_{I}(Lu_{I}(x^{(k)}))\tag{$12$}\label{eq:12}\\
&\mathbf{Y}_j^{(k)}=LSTM_{L}(Lu_{L}(y_j^{(k)}))\tag{$13$}\label{eq:13}\\
&s_j^{(k)}=\sigma(M_{2m\times 1}(\mathbf{X}^{(k)}\oplus \mathbf{Y}_j^{(k)}))\tag{$14$}\label{eq:14}\\
&l^{(k)}=-\sum_{j=1}^{C_k}\tilde{y}_j^{(k)}\log{s_j^{(k)}}\tag{$15$}\label{eq:15}
\end{align*}
where $\oplus$ denotes vector concatenation and $\tilde{y}^{(k)}$ is the true label in one-hot representation for $x^{(k)}$. The overall training objective is to minimize the weighted linear combination of costs for samples from all tasks. 

\begin{equation}
L=-\sum_{k=1}^{K}\lambda_k\sum_{i=1}^{N_k}l_{i}^{(k)}\tag{$16$}\label{eq:16}
\end{equation}
where $\lambda_k$ and $N_k$ denote the linear weight and the number of samples for each task $T_k$ as explained in Eq.(\ref{eq:4}). The network structure of the supervised model for MTLE is illustrated in Figure \ref{supervised}.

Model-\uppercase\expandafter{\romannumeral2} provides a simple and intuitive way to realize multi-task learning, where input sequences and classification labels from different tasks are jointly learned and compactly fused. During the process of training, $Lu_{I}$ and $Lu_{L}$ learn better understanding of word semantics for different tasks, while $Le_{I}$ and $Le_{L}$ obtain stronger capabilities of sequence representation. 

When new tasks are involved, it is extremely convenient for Model-\uppercase\expandafter{\romannumeral2} to scale as the whole network structure needs no modifications. We can continue training Model-\uppercase\expandafter{\romannumeral2} and further tune the parameters based on samples from the new tasks, which we define as \textbf{Hot Update}, or re-train Model-\uppercase\expandafter{\romannumeral2} again based on samples from all tasks, which is defined as \textbf{Cold Update}. We will detailedly investigate the performances of these two scaling methods in the Experiment Section.

\subsection{Model-\uppercase\expandafter{\romannumeral3}: Semi-Supervised}

For human beings, we can handle a completely new task without any more efforts and achieve appreciable performances after learning with several related tasks, which we conclude as the capability to transfer. 

\begin{figure}[!ht]
\centering
\includegraphics[width=0.45\textwidth]{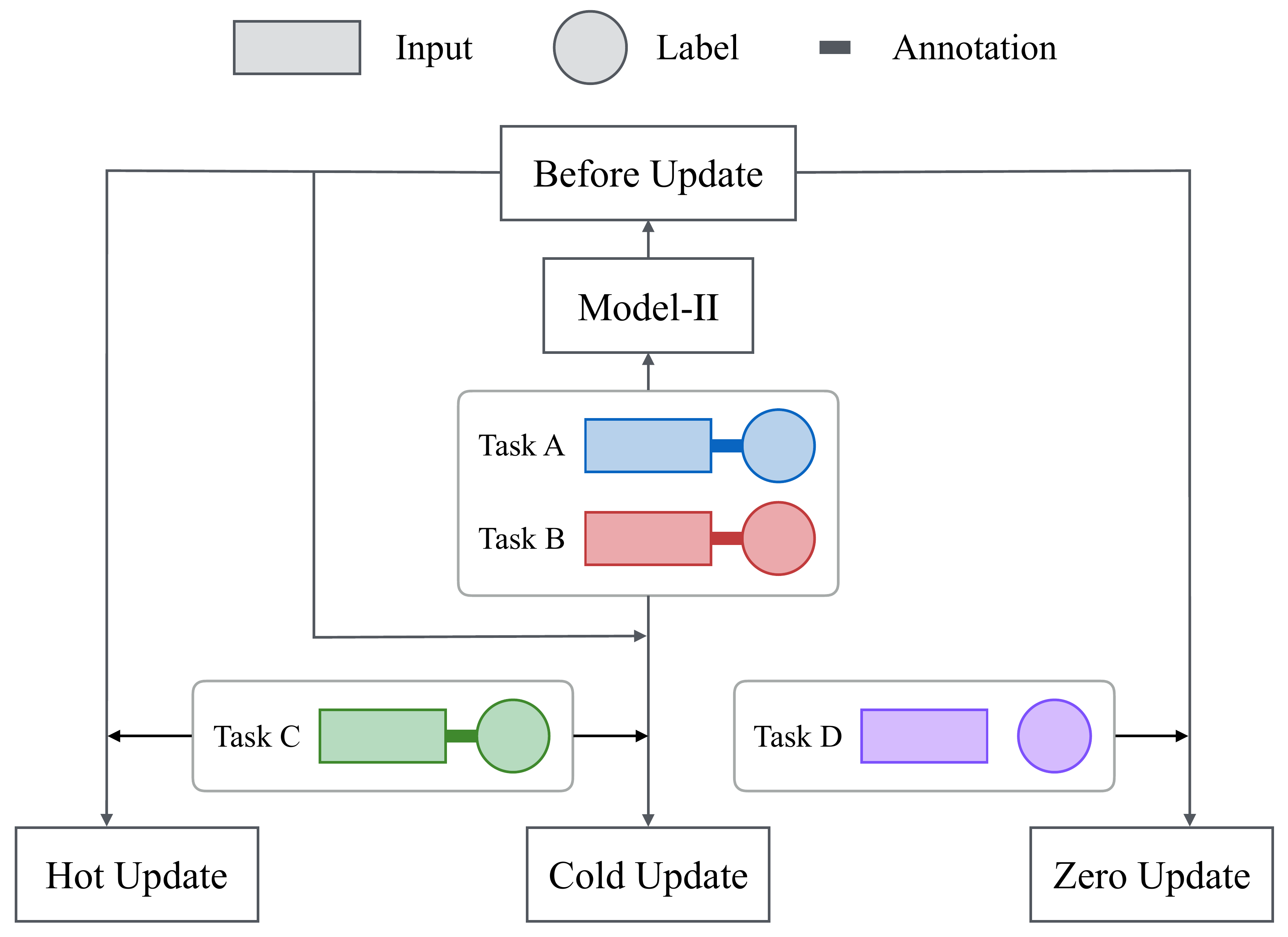}
\caption{Differences among three updating methods}\label{semi-supervised}
\end{figure}

We propose Model-\uppercase\expandafter{\romannumeral3} for semi-supervised learning based on MTLE. The only different between Model-\uppercase\expandafter{\romannumeral2} and Model-\uppercase\expandafter{\romannumeral3} is the way how they deal with new tasks, annotated or not. If the new tasks are provided with annotations, we can choose to apply Hot Update or Cold Update of Model-\uppercase\expandafter{\romannumeral2}. If the new tasks are completely unlabeled, we can still employ Model-\uppercase\expandafter{\romannumeral2} for vector mapping and find the best label for each input sequence without any further training, which we define as \textbf{Zero Update}. To avoid confusion, we specially use Model-\uppercase\expandafter{\romannumeral3} to denote the cases where annotations of new tasks are unavailable and only Zero Update is applicable, which corresponds to the transferring and semi-supervised learning capability of human beings. The differences among Hot Update, Cold Update and Zero Update are illustrated in Figure \ref{semi-supervised}, where \textbf{Before Update} denotes the model trained on the old tasks before the new tasks are introduced. We will further investigate these three updating methods in the Experiment Section.

\begin{table*}[!ht]
\centering
\caption{Five benchmark text classification datasets: SST, IMDB, MDSD, RN, QC.}\label{tab:1}
\begin{tabular}{|c|m{7cm}|c|c|c|c|c|} \hline
Dataset & Description & Type & Average Length & Class & Objective \\ \hline
SST & Movie reviews in Stanford Sentiment Treebank including SST-1 and SST-2 & Sentence & 19 / 19 & 5 / 2 & Sentiment \\ \hline
IMDB & Internet Movie Database & Document & 279 & 2 & Sentiment \\ \hline
MDSD & Product reviews on books, DVDs, electronics and kitchen appliances & Document & 176 / 189 / 115 / 97 & 2 & Sentiment \\ \hline
RN & Reuters Newswire topics classification & Document & 146 & 46 & Topics \\ \hline
QC & Question Classification & Sentence & 10 & 6 & Question Types \\ \hline
\end{tabular}
\end{table*}

\section{Experiment}

In this section, we design extensive experiments with multi-task learning based on five benchmark datasets for text classification. We investigate the empirical performances of our models and compare them to existing state-of-the-art baselines.

\subsection{Datasets}

As Table \ref{tab:1} shows, we select five benchmark datasets for text classification and design three experiment scenarios to evaluate the performances of Model-\uppercase\expandafter{\romannumeral1} and Model-\uppercase\expandafter{\romannumeral2}.

\begin{itemize}
\item \textbf{Multi-Cardinality} Movie review datasets with different average sequence lengths and class numbers, including \textbf{SST-1}~\cite{Socher-etal:2013}, \textbf{SST-2} and \textbf{IMDB}~\cite{maas-EtAl:2011:ACL-HLT2011}. 
\item \textbf{Multi-Domain} Product review datasets on different domains from \textit{Multi-Domain Sentiment Dataset}~\cite{DBLP:conf/acl/BlitzerDP07}.
\item \textbf{Multi-Objective} Text classification datasets with different objectives, including \textbf{IMDB}, \textbf{RN}~\cite{DBLP:journals/tois/ApteDW94} and \textbf{QC}~\cite{DBLP:conf/coling/LiR02}.
\end{itemize}

\subsection{Hyperparameters and Training}

Training of Model-\uppercase\expandafter{\romannumeral2} is conducted through back propagation with stochastic gradient descent~\cite{DBLP:journals/ijon/Amari93}. Besides the $E_{unsup}$ from Model-\uppercase\expandafter{\romannumeral1}, we also obtain a pre-trained lookup table by applying \textit{Word2Vec}~\cite{DBLP:journals/corr/abs-1301-3781} on the Google News corpus, which contains more than 100B words with a vocabulary size of about 3M. During each epoch, we randomly divide samples from different tasks into batches of fixed size. For each iteration, we randomly select one task and choose an untrained batch from the task, calculate the gradient and update the parameters accordingly.

All involved parameters of neural layers are randomly initialized from a truncated normal distribution with zero mean and standard deviation. We apply 10-fold cross-validation and different combinations of hyperparameters are investigated, of which the best one is described in Table \ref{tab:2}.

\begin{table}[!ht]
\centering
\caption{Hyperparameter settings}\label{tab:2}
\begin{tabular}{|c|c|} \hline
Embedding size & $d=300$ \\ \hline
Hidden layer size of LSTM & $m=100$ \\ \hline
Batch size & $\delta=32$ \\ \hline
Initial learning rate & $\eta=0.1$ \\ \hline
Regularization weight & $\lambda=10^{-5}$ \\ \hline
\end{tabular}
\end{table}

\subsection{Results of Model-\uppercase\expandafter{\romannumeral1} and Model-\uppercase\expandafter{\romannumeral2}}

We compare the performances of Model-\uppercase\expandafter{\romannumeral1} and Model-\uppercase\expandafter{\romannumeral2} with the implementation of~\cite{DBLP:journals/corr/Graves13} as shown in Table \ref{tab:3}.

It is expected that Model-\uppercase\expandafter{\romannumeral1} falls behind~\cite{DBLP:journals/corr/Graves13} as no annotations are available at all. However, with contextual information of both sequences and labels, Model-\uppercase\expandafter{\romannumeral1} still achieves considerable margins against random choices. Model-\uppercase\expandafter{\romannumeral1} performs better on tasks of shorter lengths, for example, SST-1 and SST-2, as it is difficult for unsupervised methods to learn long-term dependencies.

Model-\uppercase\expandafter{\romannumeral2} obtains significant performance gains with label information and additional correlations from related tasks. Multi-Domain, Multi-Cardinality and Multi-Objective benefit from MTLE with average improvements of 5.8\%, 3.1\% and 1.7\%, as they contain increasingly weaker relevance among tasks. The result of Model-\uppercase\expandafter{\romannumeral2} for IMDB in Multi-Cardinality is slightly better than that in Multi-Objective ~(91.3 against 90.9), as SST-1 and SST-2 share more semantically useful information with IMDB than RN and QC.

\begin{table*}[!ht]
\centering
\caption{Results of Model-\uppercase\expandafter{\romannumeral1} and Model-\uppercase\expandafter{\romannumeral2} on different scenarios}\label{tab:3}
\begin{tabular}{|c|c|c|c|c|c|c|c|c|c|c|c|}
\hline
\multirow{2}{*}{\textbf{Model}} & \multicolumn{3}{c|}{\textbf{Multi-Cardinality}} & \multicolumn{4}{c|}{\textbf{Multi-Domain}} & \multicolumn{3}{c|}{\textbf{Multi-Objective}} & \multirow{2}{*}{Avg$\Delta$} \\ \cline{2-11} 
& SST-1 & SST-2 & IMDB & Books & DVDs & Electronics & Kitchen & IMDB & RN & QC & \\ \hline
Single Task & 45.9 & 85.8 & 88.5 & 78.0 & 79.5 & 81.2 & 81.8 & 88.5 & 83.6 & 92.5 & - \\ \hline
Random & 20.0 & 50.0 & 50.0 & 50.0 & 50.0 & 50.0 & 50.0 & 50.0 & 2.2 & 16.7 & -41.6 \\ \hline
Model-\uppercase\expandafter{\romannumeral1} & 31.4 & 71.6 & 67.5 & 68.8 & 67.0 & 69.1 & 69.3 & 67.2 & 70.4 & 52.3 & -17.1 \\ \hline
Model-\uppercase\expandafter{\romannumeral2} & \textbf{49.8} & \textbf{88.4} & \textbf{91.3} & \textbf{84.5} & \textbf{85.2} & \textbf{87.3} & \textbf{86.9} & \textbf{90.9} & \textbf{85.5} & \textbf{93.2} & +3.7 \\ \hline
\end{tabular}
\end{table*}

\begin{table*}[!ht]
\centering
\caption{Results of Hot Update, Cold Update and Zero Update in different cases}\label{tab:4}
\begin{tabular}{|c|c|c|c|c|c|c|c|c|c|c|c|}
\hline
\multirow{2}{*}{\textbf{Model}} & \multicolumn{3}{c|}{\textbf{Case 1}} & \multicolumn{4}{c|}{\textbf{Case 2}} & \multicolumn{3}{c|}{\textbf{Case 3}}\\ \cline{2-11} 
& SST-1 & SST-2 & IMDB & Books & DVDs & Electronics & Kitchen & IMDB & RN & QC\\ \hline
Before Update & 48.6 & 87.6 & - & 83.7 & 84.5 & 85.9 & - & - & 84.8 & \textbf{93.4}\\ \hline
Cold Update & \textbf{49.8} & \textbf{88.3} & \textbf{91.4} & \textbf{84.4} & \textbf{85.2} & \textbf{87.2} & 86.9 & \textbf{91.0} & \textbf{85.5} & 93.2\\ \hline
Hot Update & 49.5 & 88.0 & 91.3 & 84.1 & 84.8 & 86.9 & \textbf{87.0} & 90.9 & 85.1 & 92.9\\ \hline
Zero Update & - & - & 89.9 & - & - & - & 86.3 & 74.2 & - & -\\ \hline
\end{tabular}
\end{table*}

\subsection{Scaling and Transferring Capability of MTLE}

In order to investigate the scaling and transferring capability of MTLE, we use $A+B\rightarrow C$ to denote the case where Model-\uppercase\expandafter{\romannumeral2} is trained on task $A$ and $B$, while $C$ is the newly involved one. We design three cases based on different scenarios and compare the influences of Hot Update, Cold Update, Zero Update on each task,

\begin{itemize}
\item \textbf{Case 1} SST-1 $+$ SST-2 $\rightarrow$ IMDB.
\item \textbf{Case 2} Books $+$ DVDs $+$ Electronics $\rightarrow$ Kitchen.
\item \textbf{Case 3} RN $+$ QC $\rightarrow$ IMDB. 
\end{itemize}
where in Zero Update, we ignore the training set of $C$ and directly utilize the test set for evaluations.

As Table \ref{tab:4} shows, Before Update denotes the model trained on the old tasks before the new tasks are involved, so only evaluations on the old tasks are conducted, which outperform the Single Task in Table \ref{tab:3} by 3.1\% on average.

Cold Update re-trains Model-\uppercase\expandafter{\romannumeral2} again based on both the old tasks and the new tasks, thus achieving similar performances with those of Model-\uppercase\expandafter{\romannumeral2} in Table \ref{tab:3}. Different from Cold Update, Hot Update resumes training only on the new tasks, requires much less training time, while still obtains competitive results with Cold Update. The new tasks like IMDB and Kitchen benefit more from Hot Update than the old tasks, as the parameters are further tuned according to annotations from these new tasks. Based on Cold Update and Hot Update, MTLE can easily scale and needs no structural modifications when new tasks are introduced.

Zero Update provides inspiring possibilities for completely unlabeled tasks. There are no more annotations available for additional training from the new tasks, so we can only employ the models of Before Update for evaluations on the new tasks. Zero Update achieves competitive performances in Case 1 (89.9 for IMDB) and Case 2 (86.3 for Kitchen), as tasks from these two cases all belong to sentiment datasets of different cardinalities or domains that contain rich semantic correlations with each other. However, the result for IMDB in Case 3 is only 74.2, as sentiment shares less relevance with topic classification and question type judgment, thus resulting in poor transferring performances.

\subsection{Multi-Task or Label Embedding}

MTLE mainly consists of two parts, label embedding and multi-task learning, so both implicit information from labels and potential correlations from other tasks make differences. In this section, we conduct experiments to explore the respective contributions of label embedding and multi-task learning.

We choose the four tasks from Multi-Domain scenario and train Model-\uppercase\expandafter{\romannumeral2} on each task respectively. Given that each task is trained separately, in this case their performances are only influenced by label embedding. Then we re-train Model-\uppercase\expandafter{\romannumeral2} from scratch for every two tasks, every three tasks from them and record the performances of each task in different cases, where both label embedding and multi-task learning matter. 

\begin{table*}[!ht]
\centering
\caption{Comparisons of Model-\uppercase\expandafter{\romannumeral2} against state-of-the-art models}\label{tab:5}
\begin{tabular}{|c|c|c|c|c|c|c|c|c|c|c|}
\hline
\textbf{Model} & SST-1 & SST-2 & IMDB & Books & DVDs & Electronics & Kitchen & QC \\ \hline
NBOW & 42.4 & 80.5 & 83.6 & - & - & - & - & 88.2 \\ \hline
PV & 44.6 & 82.7 & \textbf{91.7} & - & - & - & - & 91.8 \\ \hline
MT-CNN & - & - & - & 80.2 & 81.0 & 83.4 & 83.0 & - \\ \hline
MT-DNN & - & - & - & 79.7 & 80.5 & 82.5 & 82.8 & - \\ \hline
MT-RNN & 49.6 & 87.9 & 91.3 & - & - & - & - & - \\ \hline
DSM & 49.5 & 87.8 & 91.2 & 82.8 & 83.0 & 85.5 & 84.0 & - \\ \hline
GRNN & 47.5 & 85.5 & - & - & - & - & - & \textbf{93.8} \\ \hline
Model-\uppercase\expandafter{\romannumeral2} & \textbf{49.8} & \textbf{88.4} & 91.3 & \textbf{84.5} & \textbf{85.2} & \textbf{87.3} & \textbf{86.9} & 93.2 \\ \hline
\end{tabular}
\end{table*}

\begin{figure}[!ht]
\centering
\includegraphics[width=0.45\textwidth]{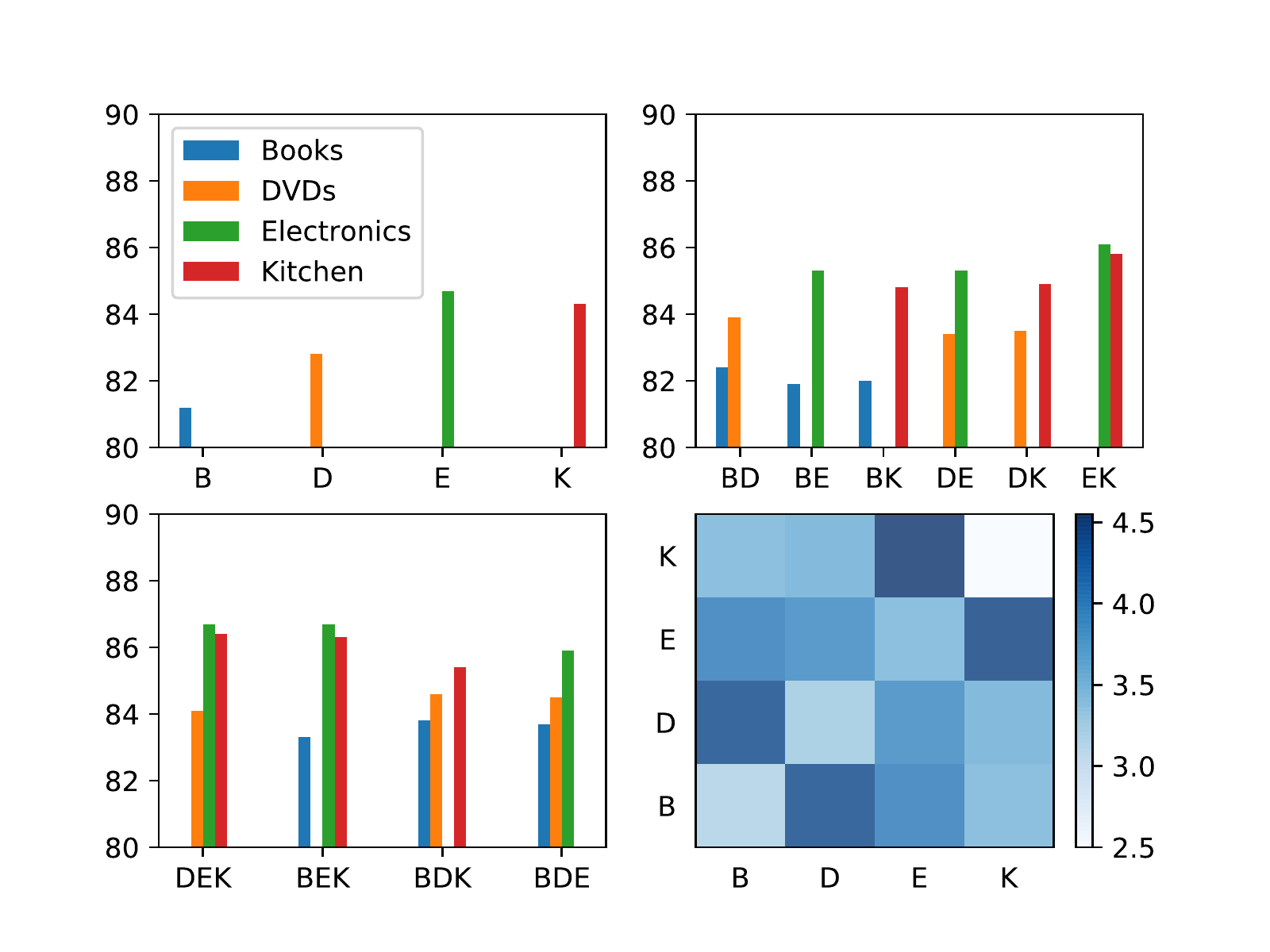}
\caption{Performance gains of each task in different cases}\label{matters}
\end{figure}

The results are illustrated in Figure \ref{matters}, where B, D, E, K are short for Books, DVDs, Electronics and Kitchen. The first three graphs denote the results of Model-\uppercase\expandafter{\romannumeral2} trained on every one task, every two tasks and every three tasks. In the first graph, the four tasks are trained separately and achieve improvements of 3.2\%, 3.3\%, 3.5\%, 2.5\% respectively compared to the baseline~\cite{DBLP:journals/corr/Graves13}. As more tasks are involved step by step, Model-\uppercase\expandafter{\romannumeral2} produces increasing performance gains for each task and achieves an average improvement of 5.9\% when all the four tasks are trained together. So it can be concluded that information from labels as well as correlations from other tasks account for considerable parts of contributions, and we integrate both of them into MTLE with the capabilities of scaling and transferring.

In the last graph, diagonal cells denote improvements of every one task, while off-diagonal cells denote average improvements of every two tasks, so an off-diagonal cell of darker color indicates stronger correlations between the corresponding two tasks. An interesting finding is that Books is more related with DVDs and Electronics is more relevant to Kitchen. A possible reason may be that Books and DVDs are products targeted for reading or watching, while customers care more about appearances and functionalities when talking about Electronics and Kitchen.

\subsection{Comparisons with State-of-the-art Models}

We compare Model-\uppercase\expandafter{\romannumeral2} against the following state-of-the-art models:

\begin{itemize}
\item \textbf{NBOW} Neural Bag-of-Words that sums up embedding vectors of all words and applies a non-linearity followed by a softmax layer.
\item \textbf{PV} Paragraph Vectors followed by logistic regression~\cite{DBLP:conf/icml/LeM14}.
\item \textbf{MT-CNN} Multi-Task learning with Convolutional Neural Networks~\cite{DBLP:conf/icml/CollobertW08} where lookup tables are partially shared.
\item \textbf{MT-DNN} Multi-Task learning with Deep Neural Networks~\cite{DBLP:conf/naacl/LiuGHDDW15} that utilizes bag-of-word representations and a hidden shared layer.
\item \textbf{MT-RNN} Multi-Task learning with Recurrent Neural Networks by a shared-layer architecture~\cite{DBLP:conf/ijcai/LiuQH16}. 
\item \textbf{DSM} Deep multi-task learning with Shared Memory~\cite{DBLP:conf/emnlp/LiuQH16} where a external memory and a reading/writing mechanism are introduced.
\item \textbf{GRNN} Gated Recursive Neural Network for sentence modeling and text classification~\cite{DBLP:conf/emnlp/ChenQZWH15}.
\end{itemize}

As Table \ref{tab:5} shows, MTLE achieves competitive or better performances on all tasks except for the task QC, as it contains less correlations with other tasks. PV slightly surpasses MTLE on IMDB (91.7 against 91.3), as sentences from IMDB are much longer than SST and MDSD, which require stronger capabilities of long-term dependency learning. In this paper, we mainly focus the idea and effects of integrating label embedding with multi-task learning, so we just apply~\cite{DBLP:journals/corr/Graves13} to realize $Le_I$ and $Le_L$, which can be further implemented by other more effective sentence learning models~\cite{DBLP:conf/emnlp/LiuQCWH15,DBLP:conf/emnlp/ChenQZWH15} and produce better performances.

\section{Related Work}

There are a large body of literatures related to multi-task learning with neural networks in NLP~\cite{DBLP:conf/icml/CollobertW08,DBLP:conf/naacl/LiuGHDDW15,DBLP:conf/emnlp/LiuQH16,DBLP:conf/ijcai/LiuQH16,ijcai2017-473}.

\cite{DBLP:conf/icml/CollobertW08} utilizes a shared lookup layer for common features, followed by task-specific layers for several traditional NLP tasks including part-of-speech tagging and semantic parsing. They use a fix-size window to solve the problem of variable-length input sequences, which can be better addressed by RNN.

\cite{DBLP:conf/naacl/LiuGHDDW15,DBLP:conf/emnlp/LiuQH16,DBLP:conf/ijcai/LiuQH16,ijcai2017-473} all investigate multi-task learning for text classification.~\cite{DBLP:conf/naacl/LiuGHDDW15} applies bag-of-word representation and information of word orders are lost.~\cite{DBLP:conf/emnlp/LiuQH16} introduces an external memory for information sharing with a reading/writing mechanism for communications.~\cite{DBLP:conf/ijcai/LiuQH16} proposes three different models for multi-task learning with RNN and~\cite{ijcai2017-473} constructs a generalized architecture for RNN based multi-task learning. However, models of these papers ignore essential information of labels and mostly can only address pair-wise interactions between two tasks. Their network structures are also fixed, thereby failing to scale or transfer when new tasks are involved.

Different from the above works, our models map labels of text classification tasks into semantic vectors and provide a more intuitive way to realize multi-task learning with the capabilities of scaling and transferring. Input sequences from three or more tasks are jointly learned together with their labels, benefitting from each other and obtaining better sequence representations.

\section{Conclusion}

In this paper, we propose Multi-Task Label Embedding to map labels of text classification tasks into semantic vectors. Based on MTLE, we implement unsupervised, supervised and semi-supervised models to facilitate multi-task learning, all utilizing semantic correlations among tasks and effectively solving the problems of scaling and transferring when new tasks are involved. We explore three different scenarios of multi-task learning and our models can improve performances of most tasks with additional related information from others in all scenarios.

In future work, we would like to explore quantifications of task correlations and generalize MTLE to address other NLP tasks, for example, sequence labeling and sequence-to-sequence learning. 


\bibliographystyle{aaai.bst}
\bibliography{aaai}

\end{document}